%% file: vidln_arxiv.tex
\documentclass[10pt,twocolumn,letterpaper]{article}

\usepackage[pagenumbers]{cvpr} %

\usepackage{graphicx}
\usepackage{amsmath}
\usepackage{amssymb}
\usepackage{booktabs}
\usepackage{adjustbox}

\usepackage{multirow}

\usepackage[pagebackref,breaklinks,colorlinks]{hyperref}

\usepackage[capitalize]{cleveref}
\crefname{section}{Sec.}{Secs.}
\Crefname{section}{Section}{Sections}
\Crefname{table}{Table}{Tables}
\crefname{table}{Tab.}{Tabs.}

\newcommand{\PAR}[1]{\vskip1pt \noindent {\bf #1~}}
\newcommand{\PARbegin}[1]{\noindent {\bf #1~}}
\newcommand{\formattedparagraph}[1]{\noindent \textbf{#1}}

\newif\ifhidecomments

\newcommand{\where}{location-output\xspace}
\newcommand{\wherecap}{Location-output\xspace}
\newcommand{\whereq}{\where question\xspace}
\newcommand{\whereqs}{\where questions\xspace}
\newcommand{\whereqssec}{\wherecap Questions\xspace}

\newcommand{\nonwhere}{text-output\xspace}
\newcommand{\nonwherecap}{Text-output\xspace}

\newcommand{\nonwhereqs}{\nonwhere questions\xspace}
\newcommand{\nonwhereqscap}{\nonwherecap questions\xspace}
\newcommand{\nonwhereqssec}{Text-Output Questions\xspace}

\newcommand{\vln}{VidLN\xspace}
\newcommand{\iln}{ImLN\xspace}
\newcommand{\vlns}{VidLNs\xspace}
\newcommand{\ilns}{ImLNs\xspace}

\newcommand{\JF}{$\mathcal{J}\&\mathcal{F}$\xspace}
\newcommand{\J}{$\mathcal{J}$\xspace}
\newcommand{\F}{$\mathcal{F}$\xspace}

\def\cvprPaperID{2716} %
\def\confName{CVPR}
\def\confYear{2023}

\begin{document}

\title{Connecting Vision and Language with Video Localized Narratives}

\author{Paul Voigtlaender \quad Soravit Changpinyo \quad Jordi Pont-Tuset \quad Radu Soricut \quad Vittorio Ferrari \\
Google Research \\
{\tt\small \{voigtlaender,schangpi,jponttuset,rsoricut,vittoferrari\}@google.com}
}

\maketitle

\input{vidln_content}

{\small
\bibliographystyle{ieee_fullname}
\bibliography{abbrev_short, paul, loco}
}

\clearpage
\twocolumn[{%
 \centering
 \LARGE \textbf{Appendix\\[1cm]}
}]

\setcounter{equation}{0}
\setcounter{figure}{0}
\setcounter{table}{0}
\setcounter{page}{1}
\setcounter{section}{0}
\makeatletter
\renewcommand{\theequation}{S\arabic{equation}}
\renewcommand{\thefigure}{S\arabic{figure}}
\renewcommand{\thetable}{S\arabic{table}}
\renewcommand{\thesection}{S\arabic{section}}

\input{supplement_content}

\end{document}

%% file: vidln_content.tex
\begin{abstract}
\vspace{-2mm}
  We propose Video Localized Narratives, a new form of multimodal video annotations connecting vision and language.
  In the original Localized Narratives \cite{PontTuset20ECCV}, annotators speak and move their mouse simultaneously on an image, thus grounding each word with a mouse trace segment. However, this is challenging on a video.
  Our new protocol empowers annotators to tell the story of a video with Localized Narratives, capturing even complex events involving multiple actors interacting with each other and with several passive objects.
  We annotated 20k videos of the OVIS, UVO, and Oops datasets, totalling 1.7M words.
  Based on this data, we also construct new benchmarks for the video narrative grounding and video question answering tasks, and provide reference results from strong baseline models. Our annotations are available at \url{https://google.github.io/video-localized-narratives/}.
\end{abstract}

\begin{figure*}
    \centering
    \includegraphics[width=\linewidth]{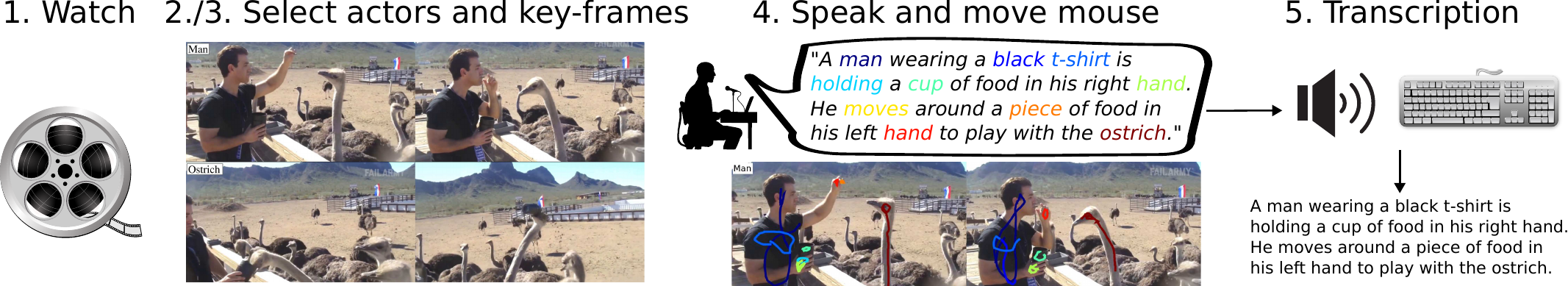}
    \caption{The five steps of the annotation process for \vlns. We are able to cleanly describe even complex interactions and events by disentangling the storylines of different actors.}
    \label{fig:annotation-process}
\end{figure*}

\begin{figure*}
    \centering
    \includegraphics[width=\linewidth]{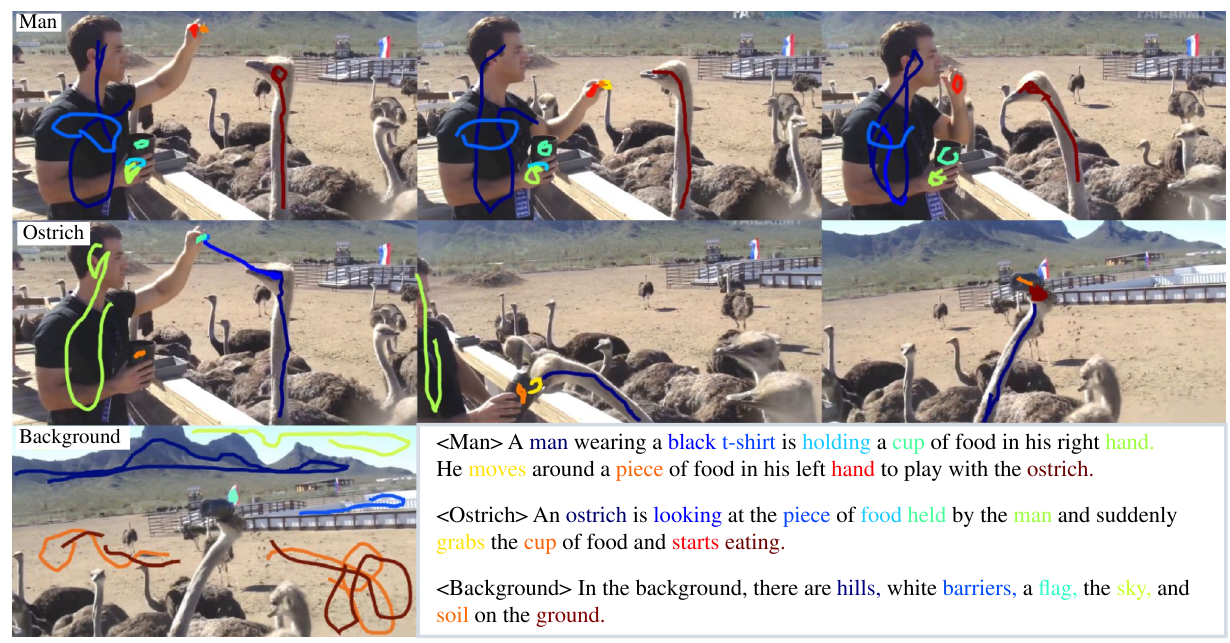}
    \caption{An example \vln annotation. Each row shows the story of the video from the perspective of a different actor. Each mouse trace segment drawn on selected key-frames localizes the word highlighted in the same color. More examples are in \cref{sec:more-vidln-examples}.}
    \label{fig:vln-example}
\end{figure*}

\vspace{-1mm}
\section{Introduction}
\vspace{-0.5em}
\label{sec:intro}
Vision and language is a very active research area, which experienced much exciting progress recently~\cite{Alayrac22flamingo,Radford21ICML_CLIP,Lu19NIPS,Ramesh21ICML,Wang21Arxiv}.
At the heart of many developments lie datasets connecting still images to captions, while grounding some of the words in the caption to regions in the image, made with a variety of protocols~\cite{chen15arxiv,young14tacl,sharma18acl,krause17cvpr,plummer17ijcv,krishna17ijcv,kazemzadeh14emnlp,mao16cvpr,PontTuset20ECCV,thapliyal2022emnlp}.
The recent Localized Narratives (\ilns)~\cite{PontTuset20ECCV} offer a particularly attractive solution: the annotators describe an image with their voice
while simultaneously moving their mouse over the regions they are describing. Speaking is natural and efficient, resulting in long captions that describe the whole scene. Moreover, the synchronization between the voice and mouse pointer yields dense visual grounding for every word. 
Yet, still images only show one instant in time. Annotating videos would be even more interesting, as they show {\em entire stories}, featuring a flow of events involving multiple actors and objects interacting with each other.

\begin{figure*}
    \centering
    \includegraphics[width=1.0\linewidth]{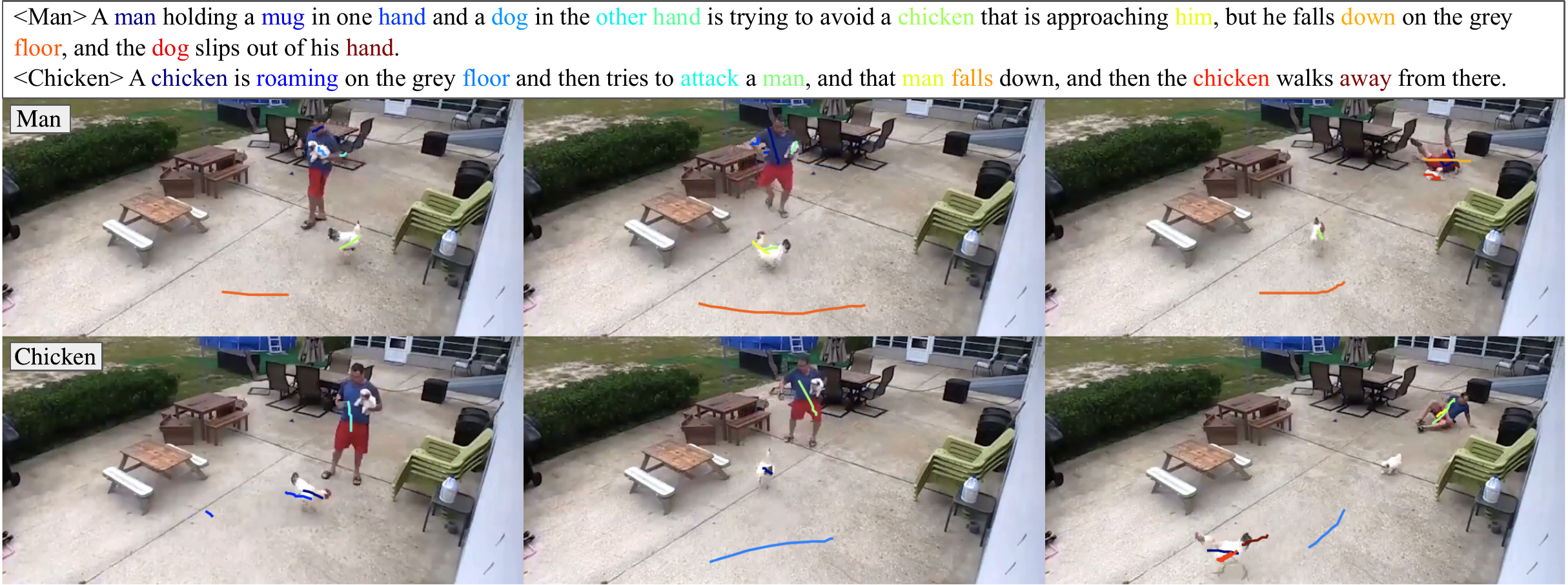}
    \caption{Another \vln example (also featuring a third actor ``Dog'', and ``Background'', not shown). From this \vln, we automatically generate text-output Q+A pairs, including ``What falls out of the man's hand?'' with answer ``dog''.
    }
    \label{fig:vln-second-example-with-where}
\end{figure*}

Directly extending \ilns to video by letting the annotator move their mouse and talk while the video is playing would lead to a ``race against time'', likely resulting in following only one salient object. In this paper, we propose a better annotation protocol which allows the annotator to tell the story of the video in a calm environment (\cref{fig:annotation-process}-\ref{fig:vln-second-example-with-where}).
The annotators first watch the video carefully, identify the main actors (``man'', ``ostrich''), and select a few representative key-frames for each actor.
Then, for each actor separately, the annotators tell a story: describing the events it is involved in using their voice, while moving the mouse on the key-frames over the objects and actions they are talking about.
The annotators mention the actor name, its attributes, and especially the actions it performs, both on other actors (\eg, ``play with the ostrich'') and on passive objects (\eg, ``grabs the cup of food'').
For completeness, the annotators also briefly describe the background in a separate step (bottom row).
Working on keyframes avoids the race against time, and producing a separate narration for each actor enables disentangling situations and thus cleanly capturing even complex events involving multiple actors interacting with each other
and with several passive objects.
As in \iln, this protocol localizes each word with a mouse trace segment.
We take several additional measures to obtain accurate localizations, beyond what was achieved in \cite{PontTuset20ECCV}.

We annotated the OVIS~\cite{Qi22IJCV_OVIS}, UVO~\cite{Wang21ICCV_UVO}, and Oops~\cite{Epstein20CVPR} datasets with Video Localized Narratives (\vlns).
These datasets cover a general domain, as opposed to only cooking~\cite{damen18eccv,Zhou18Arxiv} or first-person videos \cite{grauman2022cvpr}.
Moreover, these videos contain complex scenes with interactions between multiple actors and passive objects, leading to interesting stories that are captured by our rich annotations.
In total our annotations span 20k videos, 72k actors, and $1.7$ million words.
On average, each narrative features transcriptions for $3.5$ actors, totaling $75.1$ words, including $23.0$ nouns, $9.5$ verbs, and $8.5$ adjectives (\cref{fig:postag}).
Our analysis demonstrates the high quality of the data.
The text descriptions almost always mention objects/actions that are actually present in the video, and the mouse traces do lie inside their corresponding object in most cases.

The richness of our \vlns makes them a strong basis for several tasks.
We demonstrate this by creating new benchmarks for the tasks of Video Narrative Grounding (VNG) and Video Question Answering (VideoQA).
The new VNG task requires a method to localize the nouns in an input narrative with a segmentation mask on the video frames.
This is a complex task that goes beyond open-vocabulary semantic segmentation~\cite{ghiasi2022eccv},
as often the text has multiple identical nouns that need to be disambiguated using the context provided by other words.
We construct two VNG benchmarks on OVIS and UVO, with a total of 8k videos involving 45k objects.
We also build baseline models that explicitly attempt this disambiguation and report experiments showing it's beneficial on this dataset, while also highlighting these new benchmarks are far from solved.

For VideoQA, we construct a benchmark on the Oops dataset, consisting of \nonwhereqs (\eg, ``What is the person in the blue suit doing?''), where the answer is free-form text (\eg, ``paragliding''), and \whereqs (\eg, ``Where is the woman that is wearing a black and white dress?''), where the answer is a spatio-temporal location in the video.
The benchmark features a total of 62k questions on 9.5k videos. Answering many of these questions requires a deep understanding of the whole video.
We also implement baseline methods to establish initial results on this task.

\section{Related Work}
\label{sec:related-work}
\vspace{-0.5mm}

\PARbegin{Datasets Connecting Vision and Language.}
Many datasets exist that connect vision and language on still images, at different granularities of grounding~\cite{chen15arxiv,young14tacl,sharma18acl,krause17cvpr,plummer17ijcv,krishna17ijcv,kazemzadeh14emnlp,mao16cvpr,PontTuset20ECCV,thapliyal2022emnlp}.
In the video domain, there is also a wide range of vision-and-language datasets.
Ego4D~\cite{grauman2022cvpr} and Epic-Kitchens~\cite{damen2022ijcv} are large-scale collections of \textit{daily-life} egocentric videos accompanied with narrations, which were also used as a seed for creating benchmarks for different tasks.
We instead focus on third-person view videos that were especially selected to contain complex interactions between multiple actors and passive objects.
Our per-actor narration annotation protocol enables disentangling situations and thus cleanly describing such complex videos.
Moreover, our descriptions are long story-like narrations connecting various objects, actions, and attributes (\eg, \cref{fig:vln-example}); instead of short descriptions of atomic actions (\eg, ``C closes bottle'').
Finally, Ego4D provides grounding only for some nouns, Epic-Kitchens for all nouns, whereas our \vln annotations ground every word (including adjectives, verbs, \etc).

YouCook2-BB~\cite{Zhou18Arxiv} and ActivityNet-Entities~\cite{zhou2019cvpr} also contain video descriptions with grounded noun phrases.
YouCook2-BB focuses only on cooking, and provides grounding (bounding boxes) only for the test set.
ActivityNet-Entities annotations ground the 432 most common nouns mentioned in ActivitiyNet-Captions~\cite{krishna2017iccv} with bounding boxes.
In contrast, our \vlns ground every word with a mouse trace.
Besides, their annotation protocol is simpler: writing sentences describing major events and then finding their corresponding video segments \cite{krishna2017iccv}. It does not facilitate the disentanglement between actors or complex actor-object interactions as our protocol does.

Many other datasets~\cite{nagrani2022arxiv,yang2021iccv,monfort2021cvpr,huang2020aacl,miech2019iccv,gella2018emnlp,zhou2018aaai,krishna2017iccv,sigurdsson2016eccv,xu2016cvpr} offer textual descriptions for videos, but have no spatial localization annotation of any kind.

\PAR{Tasks Related to VNG.}
Panoptic Narrative Grounding (PNG)~\cite{Gonzalez21ICCV} creates a panoptic segmentation that grounds the nouns of an input caption describing an image.
In contrast, our proposed VNG operates on videos and focuses on concrete objects only.
\iln{}s were automatically combined with the panoptic segmentation annotations in COCO~\cite{lin14eccv,caesar18cvpr} to create PNG.
We also leveraged pre-existing segmentations, with the key difference that we manually revised and annotated the missing ones, which greatly improves the accuracy and completeness of the benchmark.

In Referring Video Object Segmentation (R-VOS)~\cite{Seo20ECCV, Khoreva18ACCV}, each referring expression is a short phrase specifically designed to identify one object.
In contrast, VNG requires taking a whole description as input and then localizing each noun, which is a more natural formulation of the task and introduces extra challenges such as co-reference resolution.
The same noun can appear multiple times in the description, referring to different objects in the video. These cases must be disambiguated based on context offered by other words.

Video Object Grounding~\cite{Zhou18Arxiv} is close to VNG as the expected output is the grounding of the noun phrases in the descriptions, but as a bounding box instead of a segmentation, and only evaluated on the top 63 recurring objects.

\PAR{Other VideoQA benchmarks.} 
There is a wide spectrum of datasets on Video Question Answering (VideoQA)~\cite{zhong2022arxiv}.
Closest to our \whereqs are the so-called \textit{Factoid VideoQA} benchmarks with open-ended answers~\cite{zhong2022arxiv}.
Many datasets contain question-answer pairs automatically derived from video descriptions~\cite{yang2022tpami,yang2021iccv,sadhu2021naacl,xu2017acmm,ye2017sigir}
without any manual checking, and thus are not suitable as a high-quality evaluation benchmark.
Besides, some of them \cite{yang2022tpami,yang2021iccv} are based on the native audio track of instructional videos, which often mentions objects/verbs that are not actually visible in the video.
Other datasets are manually curated: WildQA~\cite{castro2022coling} focuses on outdoor videos, iVQA~\cite{yang2021iccv} covers instructional videos,
ActivityNet-QA~\cite{yu2019aaai} contains various human activities, and VideoQA~\cite{zeng2017aaai} contains varied web-crawled videos.
Our \nonwhereqs are set apart from these works in that
(i) they are on videos selected to contain complex interactions between different actors/objects and
(ii) they are seeded from dense video narrations that describe the actions from the point of view of these different actors.

\begin{table*}
\adjustbox{width=0.95\linewidth,center}{
\begin{tabular}{lrrrrll}
\toprule 
{\small{}Dataset } & {\small{}\#Videos } & {\small{}\#Narratives } & {\small{}\#Actors (per narr.) } & {\small{}\#Words (per narr.) } & {\small{}Domain } & {\small{}Grounding}\tabularnewline
\midrule 
{\small{}OVIS-\vln } & {\small{}607 } & {\small{}610 } & {\small{}1,799 (2.95) } & {\small{}28,676 (47.01) } & {\small{}General } & {\small{}Every word }\tabularnewline
{\small{}UVO-\vln } & {\small{}7,588 } & {\small{}8,587 } & {\small{}25,755 (3.00) } & {\small{}549,303 (63.97) } & {\small{}General } & {\small{}Every word }\tabularnewline
{\small{}Oops-\vln } & {\small{}12,128 } & {\small{}12,894 } & {\small{}44,422 (3.45) } & {\small{}1,080,211 (83.78) } & {\small{}General } & {\small{}Every word }\tabularnewline
\midrule 
{\small{}VidLN All } & {\small{}20,323 } & {\small{}22,091 } & {\small{}71,976 (3.54) } & {\small{}1,658,190 (75.06) } & {\small{}General } & {\small{}Every word }\tabularnewline
\midrule 
{\small{}%
ActivityNet-Ent~\cite{zhou2019cvpr} } & {\small{}14,926 } & {\small{}14,926 } & {\small{}- } & {\small{}745,876 (49.97) } & {\small{}General } & {\small{}432 nouns %
}\tabularnewline
{\small{}MPII-MD~\cite{Rohrbach17CVPR} } & {\small{}94 } & {\small{}94 } & {\small{}- } & {\small{}647,814 (6,891.7) } & {\small{}Movies } & {\small{}People names%
}\tabularnewline
{\small{}YouCook2-BB~\cite{Zhou18Arxiv} } & {\small{}667 } & {\small{}667 } & {\small{}- } & {\small{}$\sim$40,500 (60.72) } & {\small{}Cooking } & {\small{}67 objects }\tabularnewline
\midrule 
{\small{}%
Ego4D-summary~\cite{grauman2022cvpr} } & {\small{}9,643 } & {\small{}18,870 } & {\small{}- } & {\small{}165,274 (8.76) } & {\small{}1st-person } & {\small{}None }\tabularnewline
{\small{}Ego4D-narration~\cite{grauman2022cvpr} } & {\small{}9,643 } & {\small{}18,870 } & {\small{}- } & {\small{}23,924,308 (1,267.85) } & {\small{}1st-person } & {\small{}Some nouns }\tabularnewline
{\small{}Epic-Kitchens~\cite{damen18eccv} } & {\small{}273 } & {\small{}273 } & {\small{}- } & {\small{}81,501 (298.54) } & {\small{}1st-p. Cooking } & {\small{}Every noun}\tabularnewline
\bottomrule
\end{tabular}
} %
\caption{\label{tab:raw-videoln-stats}Statistics of our Video Localized Narrative annotations for three datasets, compared to related datasets. Our
datasets focus on a general domain (not movies, cooking, or first-person) and provide grounding for each word.}
\end{table*}

Related to our \whereqs are the VideoQA datasets that ground parts of their textual questions and answers (\eg, noun phrases) with bounding boxes on the video~\cite{lei2019acl,choi2021aaai,yun2021iccv}.
Closest to ours is TVQA+~\cite{choi2021aaai}, where the task is to choose one answer out of five and ground it spatially and temporally, on videos from a single TV show.
In contrast, our answers can refer to any object, and our videos come from varied sources chosen to contain complex interactions.

\section{Video Localized Narrative Annotations}
A trivial extension from \ilns to video would be to simply let the annotator move their mouse and talk while the video is playing. However, this would lead to a ``race against time'' likely resulting in following only one salient object. Hence, we introduce a better protocol based on key-frames, that gives the annotator an opportunity to tell the story of the video in a calm environment (\cref{fig:annotation-process}-\ref{fig:vln-second-example-with-where}).

\subsection{Annotation Protocol}
Our annotation protocol has the following 5 steps.

\PAR{1.~Understand the Video.}
The annotator watches the video, possibly multiple times, to understand its story. %

\PAR{2.~Actor Selection.}
Some objects carry out actions, such as the man and the ostrich in \cref{fig:vln-example}, as opposed to passive objects receiving the action, such as the cup. %
The annotator selects and names the actors of this video, \eg, ``man'' and ``ostrich'' (these names are mainly a memory aid for the annotator, and not a core part of the annotations).

\PAR{3.~Key-frame Selection.}
We present the annotator with a set of key-frame candidates, uniformly sampled over time.
For each actor, the annotator selects a %
few key-frames covering the main actions it performs (the story of that actor).

\PAR{4.~A Story for each Actor.}
This is the main part of the annotation process and it is performed separately for each actor.
We show to the annotator the selected key-frames for the actor and they describe the events it is involved in, with full natural language sentences that mention the actor name, its attributes, the actions it performs, other objects and its interactions with them.
We instruct the annotators to put special emphasis on the actions performed by the actor, including also actions performed on passive objects (\eg, ``grabs the cup of food'').
While talking, the annotator moves the mouse pointer on the key-frames over the spatial positions of the objects and actions they are talking about.
For completeness, we ask the annotators to also describe the background in a separate row, albeit in less detail and typically on a single key-frame.

We want to have a good localization for every word, with a mouse trace segment.
To achieve this, we explicitly instruct the annotators to talk slowly and to stop talking while moving the mouse between two objects. Additionally, we give the annotators the option to stop a mouse trace by clicking the mouse, to avoid spurious traces, \eg, when moving the mouse between different key-frames or objects.
This results in mouse traces that are easy to segment at the beginning and end times of word utterances, leading to well-localized trace segments (see \cref{subsec:vln-statistics} for evaluation).

\PAR{5. Transcription and Time Alignment.}
After finishing steps 1-4, the annotator is asked to manually transcribe their voice recording. This ensures the captions are of high quality (as automatic speech transcription can make mistakes).

The annotations for each actor now include a mouse trace, an audio recording, and a manual transcription, but we still need an alignment between words and trace segments to provide localization.
To this end, we use an automatic algorithm \cite{Bruguier17Interspeech} to align the manual transcription directly to the audio. This yields time-stamps for each word in the transcription, revealing which part of the mouse trace belongs to which word.
Note that this direct alignment of audio to text works significantly better
than the process used by \ilns~\cite{PontTuset20ECCV} which first invokes an automatic speech-to-text transcription model, and then aligns the automatic transcriptions to the manual ones.

\subsection{Statistics}
\label{subsec:vln-statistics}

\begin{figure}[t]
    \centering
    \includegraphics[width=0.85\linewidth]{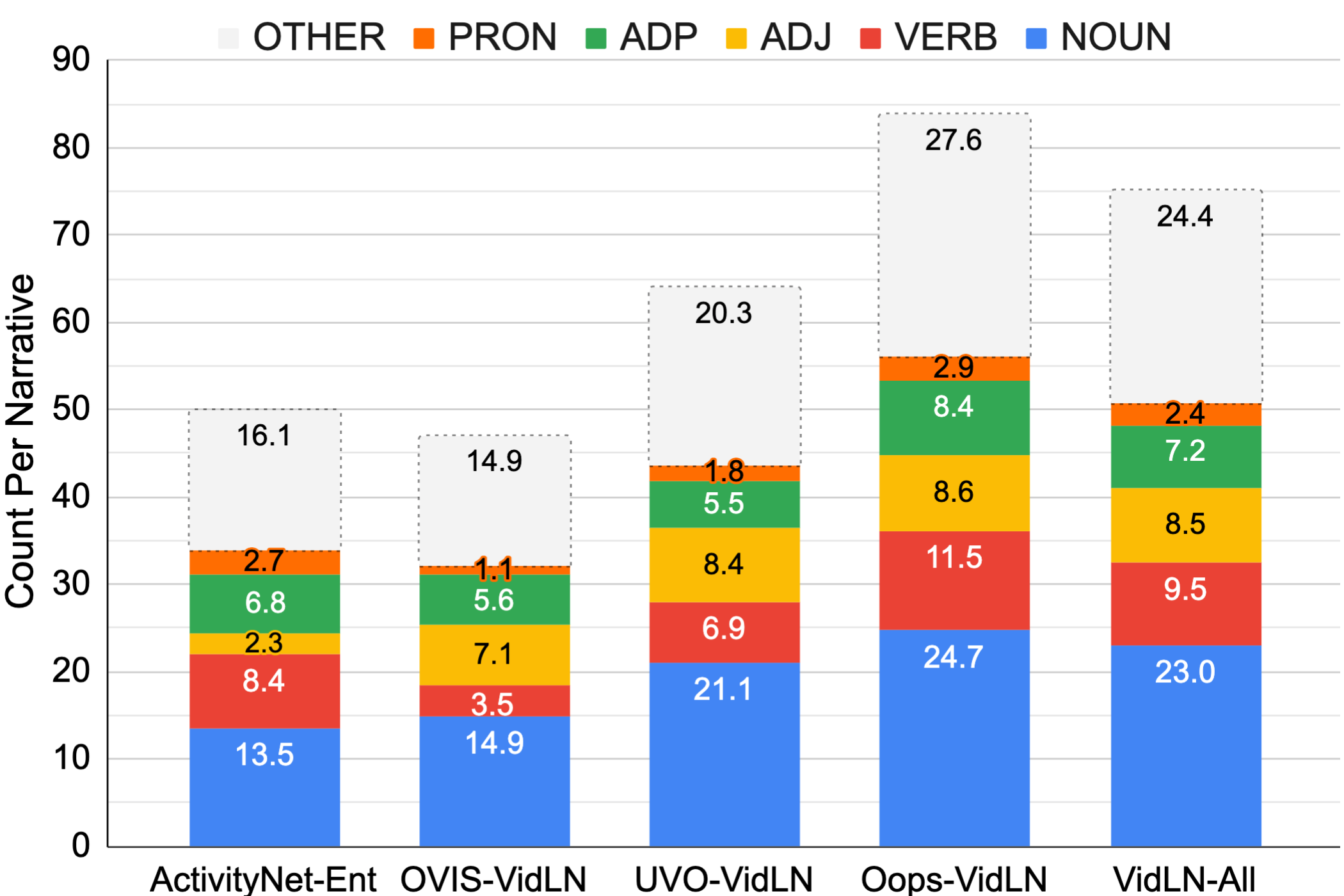}
    \caption{Richness of \vlns compared to ActivityNet-Entities. We characterize linguistic richness via word type count \emph{per narrative}. PRON: Pronouns, ADP: Adpositions, ADJ: Adjectives. Our Oops-VidLN is the most complex dataset in all major dimensions.} 
    \label{fig:postag}
    \vspace{1mm}
\end{figure}

We annotated three datasets with \vlns: OVIS \cite{Qi22IJCV_OVIS}, UVO \cite{Wang21ICCV_UVO}, and Oops \cite{Epstein20CVPR} (\cref{tab:raw-videoln-stats}).
In total, we annotated about 20k videos with over 1.65 million words, 508k of which are nouns (30.7\%), 209k verbs (12.6\%), and 188K adjectives (11.3\%). 
In contrast to existing datasets, our narratives feature a separate transcription for each actor, disentangling them (3.54 actors per video on average), and we provide a mouse trace grounding for every word. Additionally, our \vlns cover a general domain rather than only first-person~\cite{grauman2022cvpr} or cooking~\cite{damen18eccv,Zhou18Arxiv}.

In \cref{fig:postag}, we compare the per-narrative counts for the $5$ main word types to those of ActivityNet-Entities\cite{zhou2019cvpr}, the most related video dataset with grounded captions.
On average, our narratives have 75.1 words (including 23.0 nouns, 9.5 verbs, 8.5 adjectives, 7.2 adpositions, 2.4 pronouns).
Thus our narratives are longer, and contain more words of each type, than those in ActivityNet-Entities.
In summary, we present the largest and richest collection of general-domain video datasets with grounded captions to date.

\PAR{Semantic Accuracy.}
We evaluate manually how well the verbs and noun phrases of \vln describe the actions and objects in the video.
We randomly picked $70$ videos in each dataset and checked for every verb and noun phrase whether the described object/action is present in the video, and if it is correctly described.
\cref{tab:vln-semantic-correctness} shows the results, demonstrating that the captions are almost perfectly semantically accurate.

\PAR{Localization Accuracy.}
We evaluate the localization accuracy of our mouse traces on the OVIS-VNG dataset, where we have ground-truth segmentation masks for many nouns (see \cref{sec:vng}).
For each such noun we measure what percentage of its mouse trace segment is inside the ground truth mask.
We find that the average precision is high at $73.2\%$.
Additionally, we can discard mouse trace segments that consist of multiple disconnected components, as potentially indicative of some error.
Note that this can be done automatically without access to the ground-truth masks. %
This process discards $15.8\%$ of the mouse trace segments, and the remaining ones have an even higher precision of $77.3\%$.

Now we compare to the localization accuracy of the original \ilns~\cite{PontTuset20ECCV}. For \ilns, the mouse trace precision measured against ground-truth bounding boxes is $57.6\%$ for Open Images~\cite{OpenImages} and $54.8\%$ for COCO~\cite{lin14eccv} (as derived from the raw data behind Fig.~4 in~\cite{PontTuset20ECCV}, which the authors shared with us).
For our \vlns, the precision measured on bounding boxes is $83.1\%$ ($73.2\%$ measured on segmentation masks). This means that our improvements in the annotation protocol have been very successful, and that \vln traces have much higher quality than \iln traces (+25\%).

\begin{table}
\begin{centering}
\begin{tabular}{lrrr}
\toprule 
{\small{}Accuracy (\%)} & {\small{}OVIS-VLN } & {\small{}UVO-VLN } & {\small{}Oops-VLN}\tabularnewline
\midrule 
{\small{}Noun phrases } & {\small{}97.7 } & {\small{}96.8 } & {\small{}97.2}\tabularnewline
{\small{}Verbs } & {\small{}96.0 } & {\small{}97.8 } & {\small{}97.9}\tabularnewline
\bottomrule
\end{tabular}
\par\end{centering}
\caption{\label{tab:vln-semantic-correctness}
Semantic accuracy for noun phrases and verbs of our three \vln datasets.}
\end{table}

\section{Video Narrative Grounding (VNG)}
\label{sec:vng}
\begin{figure*}
    \centering
    \includegraphics[width=1.0\linewidth]{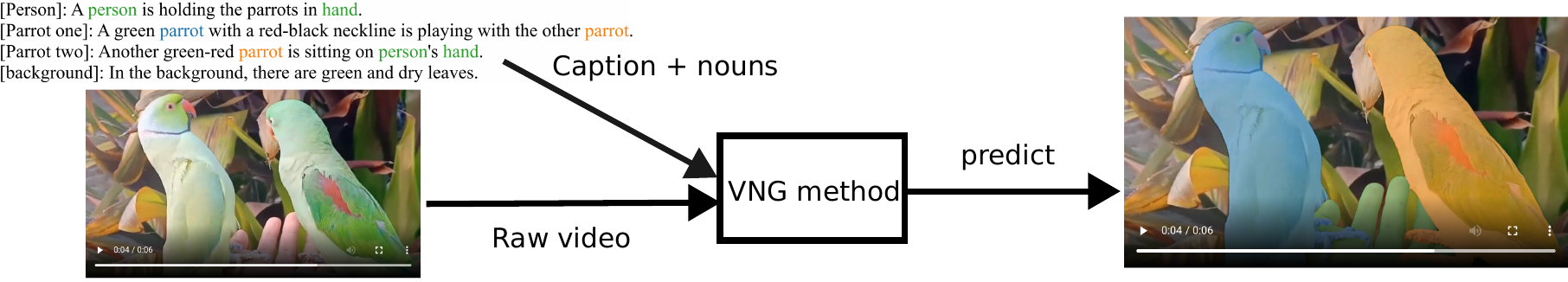}
    \caption{The Video Narrative Grounding (VNG) task. The raw video and the caption together with the position of nouns are given as input to the VNG method, which then predicts segmentation masks for each noun in each frame (masks shown in the same color as the nouns) . Note that the method has to disambiguate between two parrots based on other words, \ie, the red-black neckline.}
    \label{fig:vng-example}
\end{figure*}
\PARbegin{Task Definition.}
We propose the VNG task (\cref{fig:vng-example}), as a first example of the applicability of our \vln annotations.
The input is a video with a text description (the narrative)
and the positions of certain nouns marked.
For each marked noun, the method must output a segmentation mask for the object it refers to, in each video frame.
Importantly, the same noun (\eg, ``parrot'') can appear multiple times in the description, referring to different objects in the video. The key challenge is to disambiguate those cases, based on the context offered by other words (\eg, ``red-black neckline'').

As said in \cref{sec:related-work}, VNG is related to PNG and R-VOS, but PNG is only defined on images, and for R-VOS each referring expression is a short phrase for a single object.
Instead VNG inputs a long natural description containing multiple nouns to be disambiguated and segmented.

\PAR{Evaluation Measure.}
We adopt the well-established \JF measure \cite{DAVIS2017}, which is the average of the mean intersection-over-union \J and the boundary measure \F \cite{DAVIS2016}.
Note that different nouns can refer to the same object (\eg, ``person'' and ``hand'' in \cref{fig:vng-example}). Hence, overlapping segmentation masks are allowed and the masks for each noun are evaluated separately.

\subsection{Benchmarks}
\label{subsec:vng-benchmarks}
We propose two new benchmarks: OVIS-VNG, based on the OVIS dataset~\cite{Qi22IJCV_OVIS}, and UVO-VNG, based on the UVO dataset~\cite{Wang21ICCV_UVO}.
To build such a benchmark, we first need to select nouns in the \vln captions, and then get ground-truth masks for these nouns.
To select nouns we run a noun tagger \cite{spacy},
and then keep only singular nouns of concrete objects (\eg, car, parrot), as opposed to stuff categories (\eg, sky, water).
We then get masks for these nouns as follows.
OVIS/UVO already have mask annotations for some objects, which we ask annotators to manually match to their corresponding nouns. %
Finally, we manually annotate masks for all nouns for which OVIS/UVO does not provide one.

\begin{table}
\centering{}%
\begin{tabular}{llrr}
\toprule 
{\small{}Benchmark } & \multicolumn{1}{l}{{\small{}Task}} & \multicolumn{1}{r}{{\small{}Videos}} & {\small{}Objects (per vid.)}\tabularnewline
\midrule 
{\small{}OVIS-VNG } & {\small{}VNG } & {\small{}505 } & {\small{}2,407 (4.77)}\tabularnewline
{\small{}UVO-VNG } & {\small{}VNG } & {\small{}7,587 } & {\small{}43,058 (5.68)}\tabularnewline
\midrule 
{\small{}Ref-YTB-VOS~\cite{Seo20ECCV}} & {\small{}R-VOS } & {\small{}3,978 } & {\small{}15,009 (3.77)}\tabularnewline
{\small{}Ref-DAVIS'17~\cite{Khoreva18ACCV}} & {\small{}R-VOS } & {\small{}90 } & {\small{}205 (2.28)}\tabularnewline
\bottomrule
\end{tabular}
\vspace{1mm}
\caption{\label{tab:vng-stats} Statistics of our VNG benchmarks compared to R-VOS datasets. %
``Objects'' are not necessarily unique object instances, as 
multiple nouns or expressions can refer to the same object.
For VNG, ``Objects'' is the number of nouns with ground truth masks. %
For R-VOS, it is the number of full-video referring expressions.}
\vspace{1mm}
\end{table}

As Tab. \ref{tab:vng-stats} shows, with $7,\!587$ videos and $43,\!058$ objects, our UVO-VNG has almost $2\times$ as many videos as the largest R-VOS benchmark (Refer-YouTube-VOS), almost $3\times$ total objects, and about $1.5\times$ objects per video.
In addition to the large UVO-VNG benchmark, we propose OVIS-VNG which features many occlusions that make it challenging and its small size makes it well-suited for development.

A key challenge in VNG is to disambiguate multiple occurrences of the same noun in the input description. We analyzed the OVIS training set, which has class label annotations, and found that for $96.4\%$ of the object instances, there is another instance of the same class in the same video.
Hence, our benchmark contains substantial ambiguity that can only be resolved using context information.

\subsection{ReferFormer-VNG Baseline Method}
\label{subsec:vng-referformer-vng}
We modify the ReferFormer~\cite{Wu22CVPR} method to address our VNG task.
The original ReferFormer was designed to tackle the R-VOS task, taking a short phrase describing a single object as input.
For VNG, instead the input is a longer caption that can mention different objects and their interactions (\cref{fig:vng-example}), along with the positions of the nouns to be segmented.
For the sentence ``A green parrot with a red-black neckline is playing with the other parrot'', ReferFormer would predict a single mask per frame. For VNG, we instead need a separate mask for each parrot.

ReferFormer uses a visual encoder %
to extract features from the video. A text encoder extracts text features, from which conditional query features (conditioned on the text) are generated. 
Both types of features are fed together into a decoder to predict a mask for each frame. %
The text encoder extracts both per-text-token features and whole-sentence features. The sentence features are used to generate the conditional queries, which makes sense in the case of R-VOS, where the whole sentence describes a single object.
For ReferFormer-VNG, we instead take the features of each text-token that belongs to the noun we want to segment, and average-pool over them to obtain the conditional query (conditioned on the noun).
To segment the two different parrots, we run ReferFormer-VNG first using the features belonging to the first ``parrot'' noun, and a second time using the features of the second occurrence.

\subsection{Experiments}
\PAR{Simple baseline.}
As a simple baseline we use the original ReferFormer. We either input the full narrative as text (``Full narrative'') or we pre-process the text input based on the noun which has to be segmented (``Noun'') by keeping only the noun and add ``a'' as prefix, \eg, ``a parrot''.
The results in \cref{tab:vng-simple-baselines} show that the ``Full Narrative'' version performs poorly. This baseline inputs the same text for each noun, and hence is unable to distinguish between different nouns.
The ``Noun'' version tries to force the model to segment the correct noun, which works somewhat better. It essentially gets the class name of the object to be segmented, but cannot differentiate between different instances (\eg, two ``parrot'').
This explains the modest results, as most OVIS videos contain multiple instances of the same class (\cf \cref{subsec:vng-benchmarks}).

\PAR{ReferFormer-VNG.}
For ReferFormer-VNG's visual encoder, we use a ResNet-50 backbone \cite{resnet} initialized on ImageNet \cite{imagenet}.
For the best results, we pre-train ReferFormer-VNG on the COCO-PNG dataset which provides annotations for a similar task as VNG, but defined on images instead of videos.
Afterwards, we do the main training pass on our new UVO-VNG training set.
Finally, we can optionally fine-tune on the OVIS-VNG training set for evaluation on the OVIS-VNG test set.

\cref{tab:vng-main-results} shows results for the various training regimes.
Generally, these results are superior to those of the simple baselines, showing that ReferFormer-VNG's ability to disambiguate among multiple objects with the same class name is beneficial on this dataset.
Using the most training data (first row) leads to the strongest results. Because of having so much training data, the effect of a final fine-tuning pass on OVIS-VNG is small (from $32.0$ to $32.7$).
The second and third rows show that removing either COCO-PNG pre-training or UVO-VNG main-training reduces performance on both test sets. This also shows that training on our proposed UVO-VNG training set is beneficial, both for evaluating on similar data (the UVO-VNG test set), and for evaluating on a different dataset (OVIS-VNG), even without fine-tuning.
Implementation details and more experiments can be found in \cref{sec:referformer-vng-extra}.

\section{Video Question Answering (VideoQA)}
As a second example of the applicability of \vln annotations, we 
create a Video Question Answering (VideoQA) benchmark. We generate Q+A pairs automatically from the \vln annotations and only require humans to verify them.

Based on the \vln annotations for the Oops dataset~\cite{Epstein20CVPR}, we create Oops-QA, featuring two kinds of questions: \nonwhereqs and \whereqs (\cref{tab:vqa-stats}).
Below we describe how we generate questions, how we verify them, and present evaluation measures to assess the performance of methods for the two tasks.

\begin{table}
\begin{centering}
\begin{tabular}{lrr}
\toprule 
{\small{}Text pre-proc. } & {\small{}OVIS-VNG } & {\small{}UVO-VNG}\tabularnewline
\midrule 
{\small{}Full narrative } & {\small{}22.9 } & {\small{}25.8}\tabularnewline
{\small{}Noun } & {\small{}25.7 } & {\small{}35.6}\tabularnewline
\bottomrule
\end{tabular}
\par\end{centering}
\caption{\label{tab:vng-simple-baselines}$\mathcal{J}\&\mathcal{F}$ scores
for simple baselines for the VNG task based on the original ReferFormer.}
\vspace{1mm}
\end{table}

\subsection{\nonwhereqssec}
\nonwhereqscap are analog to those in other VideoQA benchmarks~\cite{castro2022coling,yang2021iccv,yu2019aaai,zeng2017aaai,lei2019acl,choi2021aaai,yun2021iccv},  %
where the answer is given as free-form text.
Questions can for example be about colors (``what color is the cat?''), the number of objects (``How many basketballs are in the background?''), the type of object (``The ball falls out of what?''), or about actions (``what action does the man %
perform?'').

\PAR{Automatic Q+A Generation.}
For the caption of each actor of each \vln, we use the $\mathrm{VQ}^{2}\!\mathrm{A}$ method \cite{Changpinyo22NAACL} to generate a large pool of Q+A pairs (\cref{fig:vln-second-example-with-where}).
This produces about $230$ Q+A pairs per video.
To facilitate free-form text prediction evaluation, we only keep pairs whose answer has one or two words.
Finally, to reduce redundancy we
(1) keep only one of multiple questions with the same answer, and
(2) remove near-duplicate questions.
These steps greatly reduce the number of Q+A pairs to around $22$ per video.

\PAR{Manual Verification.}
As the Q+A pairs are generated from the \vln captions that have near-perfect semantic accuracy, they are unlikely to be factually wrong. However, some Q+A pairs are irrelevant, ambiguous, or contain grammatical errors introduced by the generation algorithm.
For example: ``What color is the sky?'' is irrelevant as it can be answered without looking at the video.
The question ``What color is the cat?'' is ambiguous for a video containing two cats of different colors.
Hence, we ask two annotators to independently verify that each Q+A pair is relevant, unambiguous, and grammatically sound.
Since only verification is done manually, rather than annotating Q+A from scratch, it can be done with little effort. 
Manual verification retains $27\%$ of the generated Q+A pairs (about $6$ per video).

\begin{table}
\begin{centering}
\begin{tabular}{llrrc}
\toprule 
{\small{}COCO-PNG } & {\small{}UVO-VNG } & \multicolumn{2}{c}{{\small{}OVIS-VNG}} & {\small{}UVO-VNG}\tabularnewline
{\small{}pre-training } & {\small{}training } & {\small{}no ft } & {\small{}ft } & \tabularnewline
\midrule 
{\small{}yes } & {\small{}yes } & {\small{}32.0 } & {\small{}32.7 } & {\small{}46.4}\tabularnewline
{\small{}yes } & {\small{}no } & {\small{}28.5 } & {\small{}32.4 } & {\small{}39.6}\tabularnewline
{\small{}no } & {\small{}yes } & {\small{}26.2 } & {\small{}29.8 } & {\small{}43.1}\tabularnewline
\bottomrule
\end{tabular}
\par\end{centering}
\caption{\label{tab:vng-main-results}Results of ReferFormer-VNG on the proposed OVIS-VNG and UVO-VNG benchmarks. ``no-ft'' and ``ft''
indicate whether we fine-tuned on the OVIS-VNG training set before
evaluating on the OVIS-VNG test set. The first two columns indicate
different (pre-)training data. All numbers are $\mathcal{J}\&\mathcal{F}$
scores.}
\vspace{1mm}
\end{table}

\PAR{Evaluation Measures.}
We evaluate the text answer predicted by a VideoQA method using the exact match accuracy against the ground-truth~\cite{zhong2022arxiv,yu2019aaai}. The final result is the percentage of correct answers across the dataset.

\PAR{PaLI Baseline.}
To establish initial results on this new benchmark, we use the state-of-the-art VideoQA model PaLI~\cite{pali}.
PaLI is an encoder-decoder model that inputs an image and a text prompt, and outputs text.
We use the 1.5B version with ViT-L16~\cite{vit} (vision) and mT5-Large~\cite{mt5} (language) backbones.
At test time, the input is 1-3 video frames and the question as prompt.
Using a single frame in the middle of the video as visual input, PaLI-1.5B achieves $24.1\%$ zero-shot accuracy and $44.9\%$ when fine-tuned on the Oops-QA training set.
Using 3 equally spaced frames, the results are $25.1\%$ and $49.0\%$. %
These results demonstrate the complexity of our benchmark and leave room for the development of even more advanced techniques.

\subsection{\whereqssec}
\label{subsec:location-output-questions}
\begin{figure}
    \centering
    \includegraphics[width=1.0\linewidth]{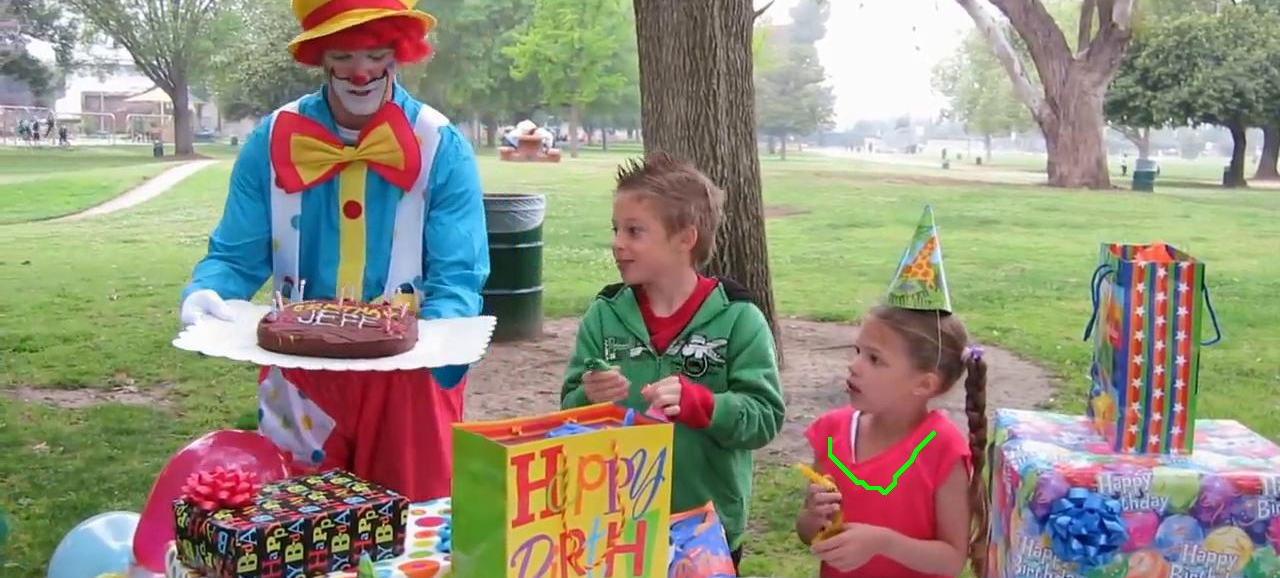}
    \caption{An example \whereq of the proposed Oops-QA benchmark: ``Where is the girl that is wearing a pink dress?'' The ground truth answer is based on a mouse trace segment (green). 
    We use a special evaluation methodology which accounts for the fact that the trace does not cover the whole object.}
    \label{fig:vqa-where-example}
    \vspace{1mm}
\end{figure}

The second part of the Oops-QA benchmark consists of \whereqs. These start with ``where is'', and their answer is a space-time location in the video (\cref{fig:vqa-where-example}).
\PAR{Automatic Q+A Generation.}
For the caption of each actor of a \vln, we first use spaCy~\cite{spacy} to assign part-of-speech tags for each word (\eg, verb, adjective), and a parse tree for each sentence (\eg, connecting subjects to their verbs). Based on this information, we try to transform each sentence into a question about where the subject is. As sentences can be long, we first use the parse tree to focus on the sub-tree containing the verb and the subject (\eg, ``the girl that is wearing a pink dress'' in a much longer caption). In this case we prepend ``where is'' to obtain the question. In other cases, we add ``that'' or ``that is'' before the verb (\eg, ``where is the dog \textit{that is} playing with the cat'').
In this manner we generate about $3.7$ questions per video.

For each question, we use the mouse trace segment associated to the subject as a basis to construct a ground-truth answer.
To facilitate the subsequent manual verification process, when the trace segment spans multiple frames, we retain only the frame with the longest segment.
To improve quality, we only consider trace segments that consist of a single connected component (\cf~\cref{subsec:vln-statistics}).

\PAR{Manual Verification.}
We show to two annotators each question together with its trace segment overlaid on the frame selected above.
We only keep a Q+A pair if both annotators agree the question is grammatically correct, the trace segment is on the correct object, and the question refers to a unique object in that video.
For example, in \cref{fig:vqa-where-example}, ``where is the kid?'' is ambiguous, whereas ``where is the kid with the pink dress'' is not.
About $52\%$ of the generated Q+A pairs pass this verification step, yielding about $2$ per video for the final benchmark.
As a side benefit, the localization accuracy of the mouse traces in the verified questions is extremely high ($92.9$\% on the correct object mask on average, see \cref{subsec:mouse-trace-precision} for details).

\PAR{Evaluation Measures.}
For each question, a VideoQA method must output a bounding box for every frame where the object is present. However, we evaluate only in the one frame where we have a verified ground-truth trace segment $t$.
For a predicted box $b$ to be correct, it has to fulfill both the precision and the recall criterion described below.

{\it Recall criterion}:
$b$ has to contain most of $t$, according to intersection-over-area: $\frac{|b\ \cap\ t|}{|t|} \geq 0.5.$

{\it Precision criterion}:
A mouse trace segment only covers part of an object surface (\cref{fig:vqa-where-example}).
To mitigate this effect, we derive from $t$ an approximate ground-truth box $g$ around the object, and use it to evaluate the precision of $b$ (see \cref{subsec:square-estimation} for how we learn a transformation to enlarge the trace segment into an approximate box around the object).
This requires most of $b$ to be contained in $g$: $\frac{|b\ \cap\ g|}{|b|} \geq 0.5$.

\PAR{Quality of Evaluation Measure.}
To demonstrate the validity of our evaluation procedure based on approximate ground-truth boxes, we perform an experiment on OVIS-VNG.
We simulate a perfect VideoQA method which outputs a bounding box on the ground-truth segmentation masks of the OVIS-VNG test set. We now evaluate these perfect predictions against our approximate ground-truth boxes.
The result is that $99.1$\% of the perfectly predicted boxes fulfill the recall criterion above, $85.1$\% fulfill the precision criterion, and $84.4\%$ fulfill both.
Hence, a strong method will be properly rewarded by our evaluation.
Additionally, on average $82.3\%$ of the image area is not covered by the approximate ground-truth box. Hence, the task is challenging: as the ground-truth boxes are small on average, a method cannot easily localize them (predictions falling outside the ground-truth are penalized by a low precision).

\PAR{Baseline Method.}
We repurpose ReferFormer-VNG (\cref{subsec:vng-referformer-vng}) to answer \whereqs.
We turn the question into a statement by dropping the ``where is'', then input it together with the position of the first noun into ReferFormer-VNG to produce a segmentation mask in each video frame.
Finally, we convert the masks into bounding boxes, and evaluate them as described above.
This baseline fulfills the recall criterion for $66.7\%$ of the questions, 
the precision criterion for $53.9\%$, and both for $48.3\%$.
While these results are good, they are far from perfect, showing the challenge posed by our new benchmark and leaving plenty of room for the development of better methods.

\begin{table}[t]
\begin{centering}
\begin{tabular}{lrrrr}
\toprule 
\multirow{1}{*}{{\small{}Oops-QA}} & \multicolumn{2}{c}{{\small{}\nonwherecap}} & \multicolumn{2}{c}{{\small{}\wherecap}}\tabularnewline
 & {\small{}train } & {\small{}test } & {\small{}train } & {\small{}test}\tabularnewline
\midrule 
{\small{}\#Videos } & {\small{}7,509 } & {\small{}1,982 } & {\small{}8,113 } & {\small{}1,691}\tabularnewline
{\small{}\#Questions (total) } & {\small{}31,760 } & {\small{}12,417 } & {\small{}14,779 } & {\small{}3,179}\tabularnewline
{\small{}\#Questions (per vid.) } & {\small{}4.23 } & {\small{}6.26 } & {\small{}1.82 } & {\small{}1.88}\tabularnewline
\bottomrule
\end{tabular}
\par\end{centering}
\caption{\label{tab:vqa-stats}Statistics of our Oops-QA benchmark, divided into \nonwhereqs and \whereqs.}
\end{table}

\subsection{Combined Score}
\vspace{-0.5em}
We define a combined score for the whole Oops-QA benchmark as the average of the two sub-tasks. Hence our two baselines together achieve an overall score of $50.8\%$ accuracy (mean of $53.2\%$ and $48.3\%$).
With the unified Oops-QA benchmark of \whereqs and \nonwhereqs, we want to encourage the development of a single model that can answer both kinds of questions and at the same time improve the results.

\vspace{-0.5em}
\section{Conclusion}
\vspace{-0.5em}
We introduce \vlns, an annotation procedure that obtains rich video descriptions, that are semantically correct and densely grounded with accurate spatio-temporal localizations. We annotated roughly 20k videos of three different datasets, and obtained captions with more than 1.6 million words in total. We demonstrated the versatility of \vlns by using them to generate two Video Narrative Grounding benchmarks, and the Oops-QA benchmark.

\formattedparagraph{\footnotesize Acknowledgement.} {\footnotesize {We would like to thank Jasper Uijlings and Thomas Mensink for helpful discussions.}}

%% file: supplement_content.tex
\section{More Video Localized Narratives Examples}
\label{sec:more-vidln-examples}
\begin{figure*}
    \centering
    \includegraphics[width=\linewidth]{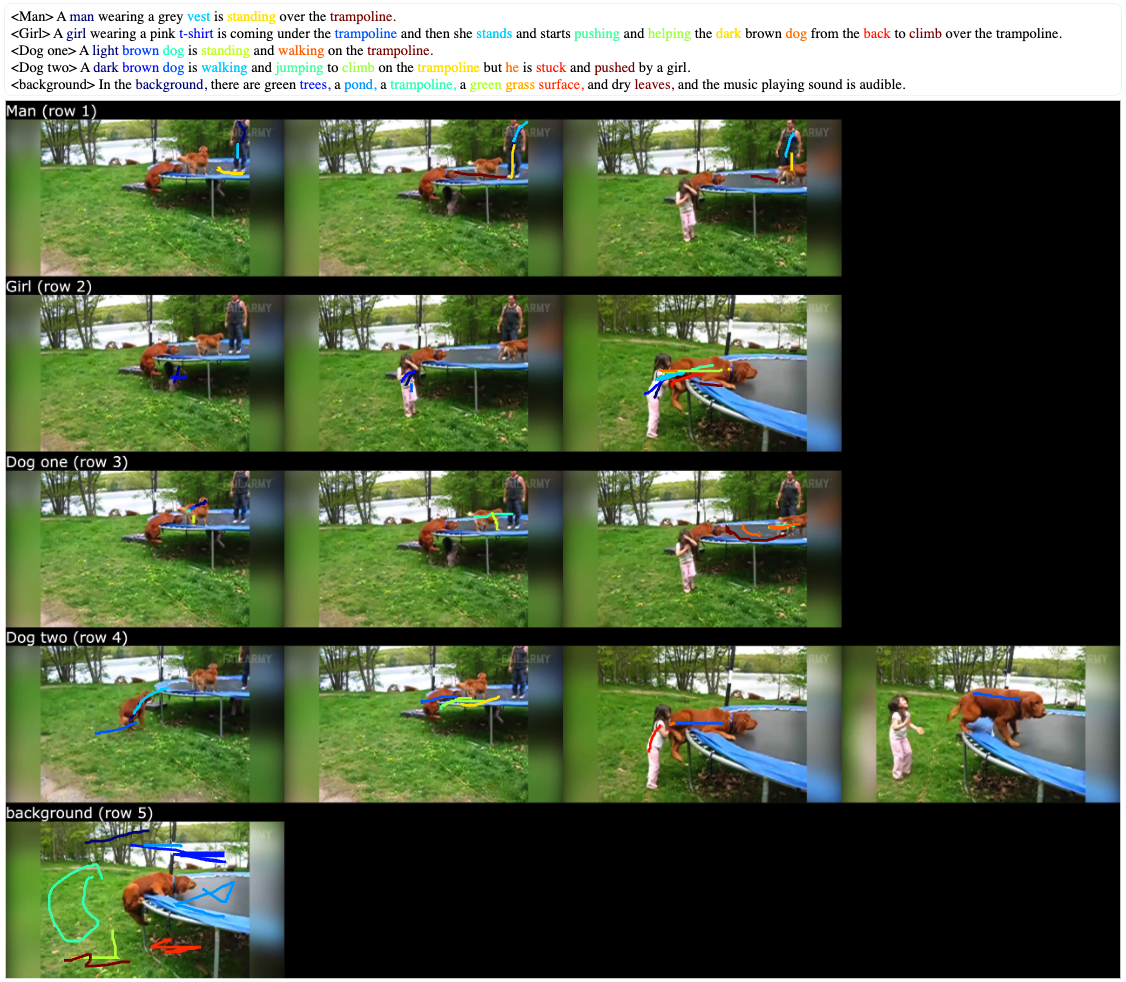}
    \caption{A \vln example with four actors and background.}
    \label{fig:vln-example-dog-trampoline}
\end{figure*}

\begin{figure*}
    \centering
    \includegraphics[width=\linewidth]{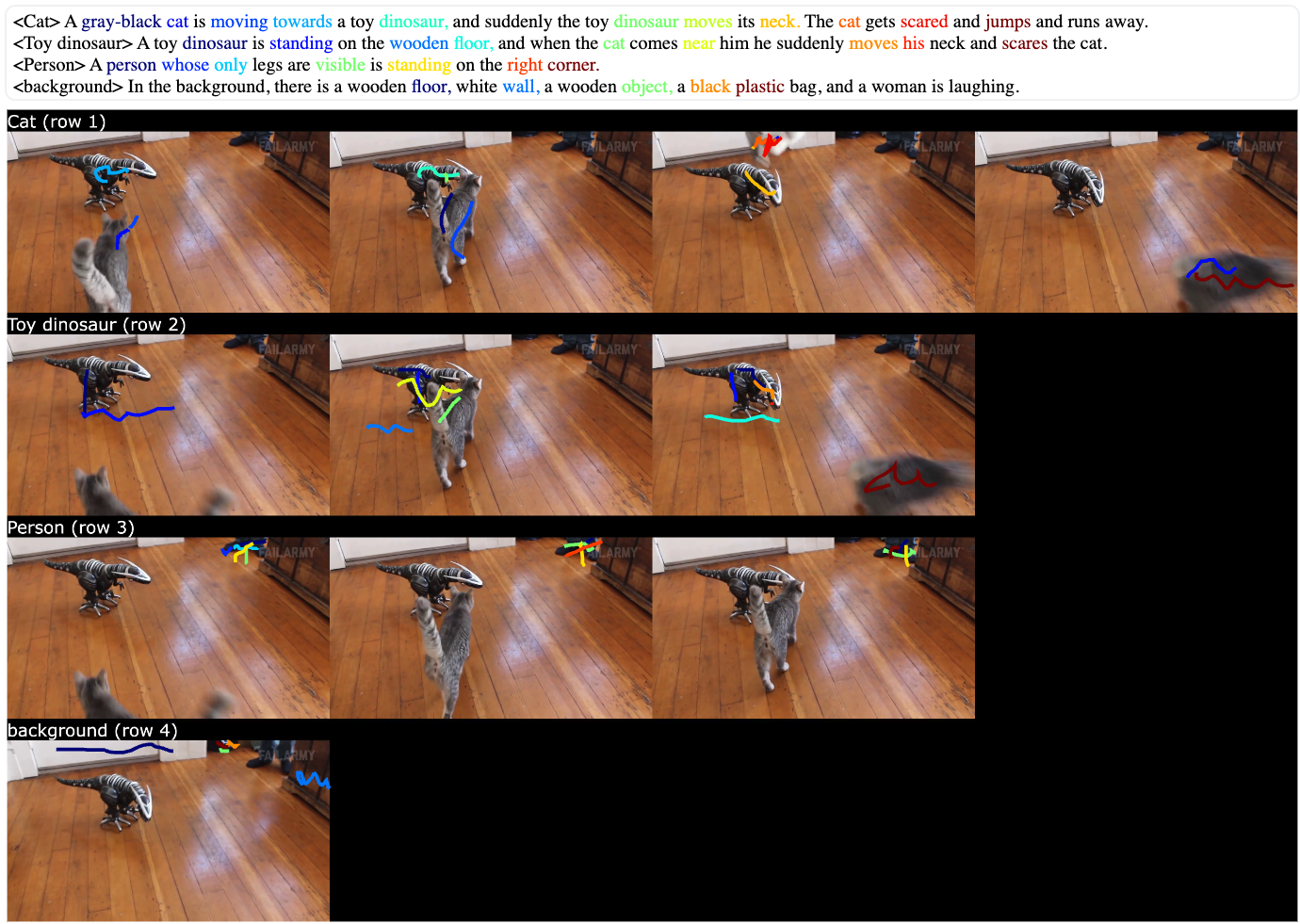}
    \caption{Another example \vln annotation.}
    \label{fig:vln-example-cat-robot}
\end{figure*}

\begin{figure*}
    \centering
    \includegraphics[width=\linewidth]{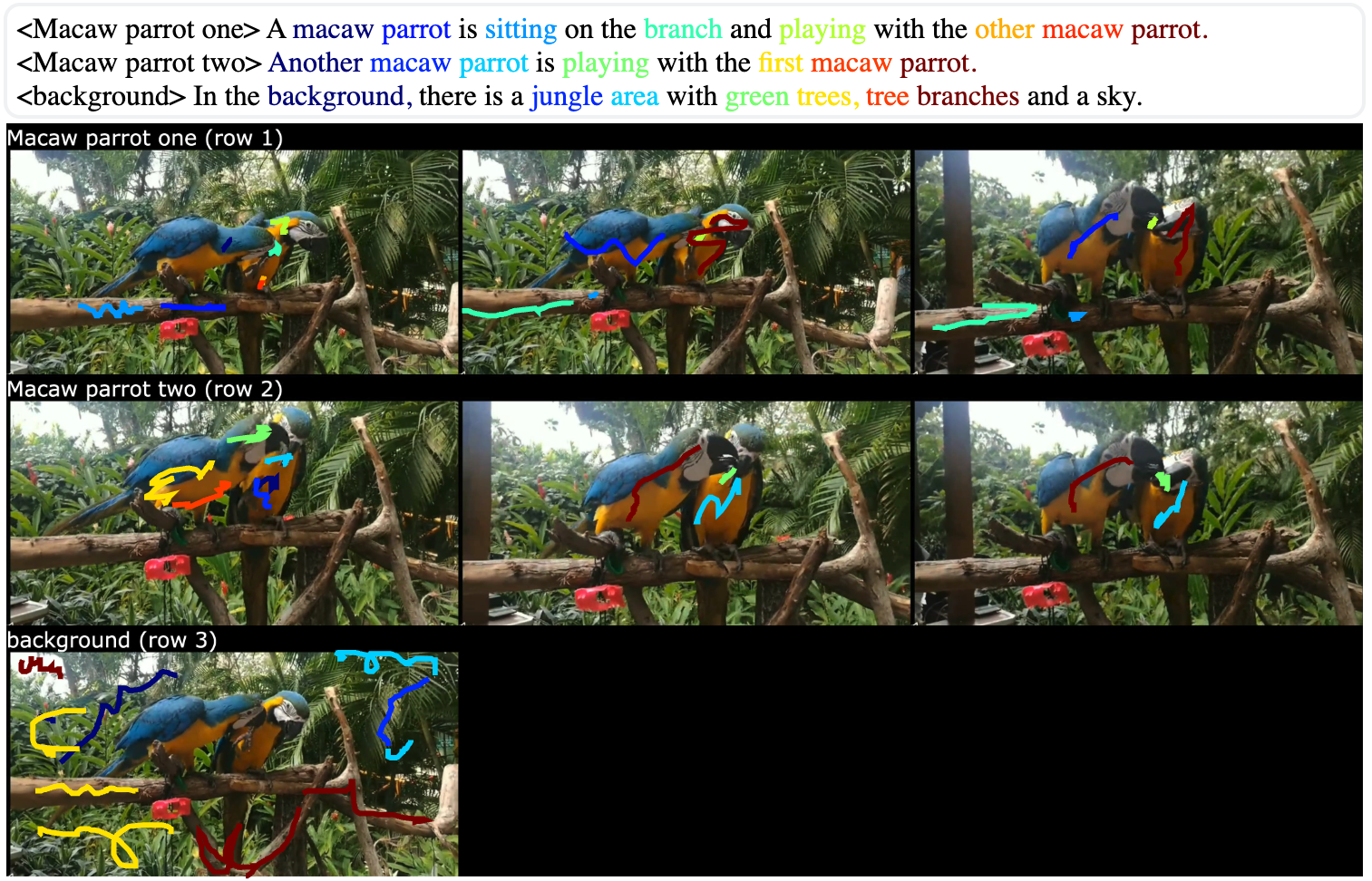}
    \caption{Another example \vln annotation.}
    \label{fig:vln-example-parrots}
\end{figure*}

In \cref{fig:vln-example-dog-trampoline,fig:vln-example-cat-robot,fig:vln-example-parrots}, we show more examples of our \vln annotations. %

\section{ReferFormer-VNG}
\label{sec:referformer-vng-extra}

Here we describe implementation details and additional experiments for ReferFormer-VNG. %

\subsection{Implementation Details}
For simplicity, for the ReferFormer-VNG baseline, we input only the narrative of one actor at a time (\eg, only the narrative of ``Parrot one'' rather than the concatenation of all narratives of all actors), and the segmentation for each noun is predicted separately, both during training and inference. During training, for videos with sparse mask annotations, we only sample from the frames that have a mask annotated.
For the text encoder, we use a pre-trained RoBERTa \cite{Liu19Arxiv} model to extract both per-text-token features and whole-sentence features.

For the training hyper-parameters, we generally follow the setup of the original ReferFormer~\cite{Wu22CVPR}, and in the following we will only describe settings that are different.
For fine-tuning on OVIS-VNG, we train for 6 epochs with 8 Tesla V100 GPUs and reduce the learning rate at epochs 3 and 5. For all other training setups, we train for 12 epochs with 16 Tesla V100 GPUs, and reduce the learning rate at epochs 8 and 10.

\subsection{Additional Experiments}

\paragraph{Experiments with Different Backbones.}
\begin{table*}
\begin{centering}
\begin{tabular}{ccccccc}
\toprule 
\multirow{2}{*}{Backbone} & Initialization & COCO-PNG & UVO-VNG & \multicolumn{2}{c}{OVIS-VNG} & \multirow{2}{*}{UVO-VNG}\tabularnewline
 & Dataset & Pre-training & Training & no ft & ft & \tabularnewline
\midrule
\multirow{2}{*}{Swin-Large} & \multirow{2}{*}{ImageNet-21k} & yes & yes & 30.7 & 36.7 & 42.8\tabularnewline
 &  & yes & no & 36.9 & 40.0 & 55.1\tabularnewline
\midrule
\multirow{4}{*}{VideoSwin} & \multirow{2}{*}{ImageNet-1k} & yes & yes & 25.4 & 28.3 & 35.9\tabularnewline
 &  & yes & no & 26.9 & 28.5 & 41.4\tabularnewline
\cmidrule{2-7} \cmidrule{3-7} \cmidrule{4-7} \cmidrule{5-7} \cmidrule{6-7} \cmidrule{7-7} 
 & \multirow{2}{*}{Kinetics{*}} & yes & yes & 31.5 & 35.9 & 42.9\tabularnewline
 &  & yes & no & 
35.5 & 38.2
 & 
53.5\tabularnewline
\midrule
\multirow{2}{*}{ResNet-50} & \multirow{2}{*}{ImageNet-1k} & yes & yes & 32.0 & 32.7 & 46.4\tabularnewline
 &  & yes & no & 28.5 & 32.4 & 39.6\tabularnewline
\bottomrule
\end{tabular}
\par\end{centering}
\caption{\label{tab:vng-other-backbones}VNG results with other visual backbones.
All numbers are $\mathcal{J}\&\mathcal{F}$ scores. The ResNet-50
result is shown again for comparison. ``no-ft'' and
``ft'' indicate whether we fine-tuned on the OVIS-VNG training set
before evaluating on the OVIS-VNG test set. {*}: Note that the videos
of the UVO-VNG test set overlap with the Kinetics training set, but
the task is different.}
\end{table*}
\cref{tab:vng-other-backbones} shows results for ReferFormer-VNG with different visual backbones. Here we use the same backbones that are considered for the original ReferFormer~\cite{Wu22CVPR}.

The order of training is always (i) initialize the visual backbone with the checkpoint obtained on the dataset specified in the column ``Initialization Dataset'', (ii) pre-train for 12 epochs on COCO-PNG, (iii) optionally train for 12 epochs on UVO-VNG, and (iv) optionally fine-tune for 6 epochs on OVIS-VNG. Here, the step (iv) is only optionally used for evaluation on OVIS-VNG.

The results for ResNet-50 are the same as in \cref{tab:vng-main-results} and are shown here again for comparison.

Using a large Swin transformer~\cite{Liu21ICCV} backbone (``Swin-Large'') pre-trained on the larger ImageNet-21k version achieves the strongest results for all setups, reaching $40.0$ \JF on OVIS-VNG (with fine-tuning) and $55.1$ on UVO-VNG.

Using instead a Video Swin Transformer~\cite{Liu22CVPR} (``VideoSwin''), for most setups also achieves stronger results than the ResNet-50 backbone. However, when initializing from ImageNet-1k and not training on UVO-VNG, then no video data (or only for OVIS-VNG fine-tuning) is used. In this case, the Video Swin Transformer which is specifically designed for videos, cannot unfold its full potential and yields relatively weak results. When initializing from Kinetics, and using UVO-VNG for main-training with videos, the VideoSwin results are strong, close to the results of Swin-Large. 

The Video Swin Transformer result with initialization on Kinetics and evaluation on UVO-VNG needs to be taken with a grain of salt, because the Kinetics~\cite{Kay17Arxiv} training dataset overlaps with the UVO-VNG test set. However, the task for Kinetics is action classification, which is very different from VNG.

\paragraph{Fine-tuning on OVIS-VNG from ImageNet.}
We perform an additional experiment, where we initialize ReferFormer-VNG's ResNet-50 backbone on ImageNet and then directly train on the OVIS-VNG training set and afterwards evaluate on the OVIS-VNG test set. This setup achieves a \JF score of $25.1$, which is much lower than the scores of $32.4$ with COCO-PNG pre-training and $32.7$ with COCO-PNG pre-training and UVO-VNG main training (see~\cref{tab:vng-other-backbones}).
This result shows that without any other VNG-related training data, the result greatly suffers.

\section{Video Question Answering: Location-output Questions}
Here we explain for the VideoQA location-output questions, how the approximate square bounding boxes are estimated, and how we estimated the precision of the mouse trace answers.

\subsection{Estimation of Approximate Square Bounding Boxes}
\label{subsec:square-estimation}
For the evaluation of location-output questions, we start from a location ground truth in the form of a annotator-verified mouse trace on the object of interest. In order to evaluate the precision criterion of predicted bounding boxes, we estimate approximate square bounding boxes based on the mouse trace segment as follows.

A mouse trace segment only has a notion of a single scale dimension (its length), not two. Hence, we estimate a squared box rather than the usual rectangular one.
\begin{figure}
    \centering
    \includegraphics[width=1.0\linewidth]{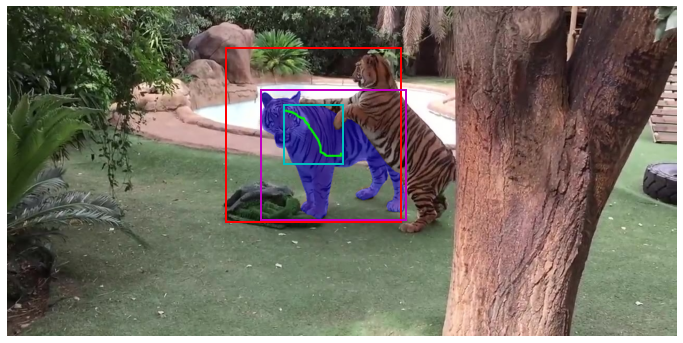}
    \caption{The estimated square bounding box (red) is obtained by scaling the (square) trace-box (cyan) of the mouse trace (green). The (unknown) ground truth mask is overlaid in blue and the (rectangular) ground truth bounding box is shown in magenta.}
    \label{fig:bounding-square-estimation-example}
\end{figure}
We start by first fitting a square around the mouse trace segment (the trace-box, cyan in \cref{fig:bounding-square-estimation-example}) by setting its center to the center of mass of the trace and setting its side-length to the smallest value with which the whole mouse trace segment is covered by the box.
Because the mouse trace only covers part of the object, this trace-box will likely be too small. Hence, to obtain the approximate bounding box we enlarge the side length by a learned transformation.

We learn a quadratic function that performs this transformation on the OVIS-VNG train dataset, as it has ground truth segmentation masks.
The side length $x$ of the trace-box is the input to the transformation, and the side length of the perfect square bounding box around the ground truth segmentation mask is the target output $\hat{y}$.

For each noun that has both an associated ground truth segmentation mask, and a mouse trace, we determine both the input side length $x$ and the desired output side length $\hat{y}$.
Our experiments revealed that the relationship between $x$ and $\hat{y}$ is best captured with a quadratic transformation, as larger $x$ need to be enlarged by a smaller factors than small $x$ to reach their corresponding $\hat{y}$:

\begin{equation}
  f(x)=\lambda_0 + \lambda_1 x + \lambda_2 x^2
\end{equation}
with three free parameters $\lambda_0, \lambda_1, \lambda_2 \in \mathbb{R}$.

As a loss function for a single training example, we consider the ratio of the estimated side length $f(x)$ to the ground-truth one $\hat{y}$. This is a scale-invariant measure, and thus more suitable for the task than an L2 loss:

\begin{equation}
  L(x,\hat{y}) = \max \left\{ \frac{f(x)}{\hat{y}}, \frac{\hat{y}}{f(x)} \right\}.
\end{equation}

Intuitively, the ratio of the predicted side length and the desired side length should be as close to $1$ as possible. The maximum is necessary to account both for cases where the prediction is too small and where it is too big. The theoretical minimum of this loss is $1.0$.

Finally, we take the geometric mean over the whole training set with $N$ inputs $x_1,\dots,x_N$, and desired outputs $\hat{y}_1,\dots,\hat{y}_N$,  \ie,
\begin{equation}
  L_{total} = \left(\prod_{n=1}^N L(x_n, \hat{y}_n)\right)^{\frac{1}{N}}.
\end{equation}

We optimize this loss on the OVIS-VNG training set with batch gradient descent, \ie, in each step the loss is calculated over the whole dataset.

\subsection{Precision of Mouse Trace Answers}
\label{subsec:mouse-trace-precision}
For the Oops dataset, we do not annotate any segmentation masks. Hence, we instead use the OVIS-VNG dataset as a proxy to measure mouse trace precision for the location-output questions of \cref{subsec:location-output-questions}.
Recall from \cref{subsec:vln-statistics} that the trace precision, without manual verification, on OVIS-VNG is $77.3\%$, \ie, $77.3\%$ of the trace points are on the correct object mask on average.
However, in \cref{subsec:location-output-questions} the candidate location-output questions are manually verified by two annotators to ensure the trace segments are correct (among other criteria). This increases the average accuracy of the questions that pass verification.
To emulate this verification process on OVIS-VNG (instead of Oops), we discard trace segments with a precision of less than $25\%$. This filtering step removes roughly $9\%$ of the mouse trace segments and the average precision of the remaining trace segments is very high at $92.9$\%. Note that the threshold was chosen conservatively and our manual verification by two annotators would likely lead to even higher precision.